\documentclass[journal]{IEEEtran}
\usepackage{graphicx}
\usepackage{comment}
\usepackage{amsmath,amssymb} 
\usepackage{color}
\usepackage{multirow}
\usepackage{subfigure}
\usepackage[colorlinks]{hyperref}
\usepackage[table,xcdraw]{xcolor}

\usepackage{booktabs}
\ifCLASSINFOpdf
\else
\fi
\begin{document}
%
\title{B$\boldsymbol{A}^2$M: A Batch Aware Attention Module for Image Classification}
%
%
%
\author{Qishang~Cheng,
	Hongliang~Li, ~\IEEEmembership{Senior~Member,~IEEE},
	Qingbo~Wu, ~\IEEEmembership{Member,~IEEE},
	and King~Ngi~Ngan,~\IEEEmembership{Fellow,~IEEE}%

	\thanks{Q. Cheng, H. Li, Q. Wu and King. N. Ngan are with the School of Electronic Engineering, University of Electronic Science and Technology of China, Chengdu  611731, China (e-mail:  cqs@std.uestc.edu.cn;  hlli@uestc.edu.cn; qbwu@uestc.edu.cn; knngan@uestc.edu.cn).}
}

\markboth{ }%
{Shell \MakeLowercase{\textit{et al.}}: B$\boldsymbol{A}^2$M: A Batch Aware Attention Module for Image Classification}
%



\maketitle

\begin{abstract}
    The attention mechanisms have been employed in Convolutional Neural Network (CNN) to enhance the feature representation. However, existing attention mechanisms only concentrate on refining the features inside each sample and neglect the discrimination between different samples. In this paper, we propose a batch aware attention module (B$\boldsymbol{A}^2$M) for feature enrichment from a distinctive perspective. More specifically, we first get the sample-wise attention representation (SAR) by fusing the channel, local spatial and global spatial attention maps within each sample. Then, we feed the SARs of the whole batch to a normalization function to get the weights for each sample. The weights serve to distinguish the features' importance between samples in a training batch with different complexity of content. The B$\boldsymbol{A}^2$M could be embedded into different parts of CNN and optimized with the network in an end-to-end manner. The design of B$\boldsymbol{A}^2$M is lightweight with few extra parameters and calculations. We validate B$\boldsymbol{A}^2$M through extensive experiments on CIFAR-100 and ImageNet-1K for the image recognition task. The results show that B$\boldsymbol{A}^2$M can boost the performance of various network architectures and outperforms many classical attention methods. Besides, B$\boldsymbol{A}^2$M exceeds traditional methods of re-weighting samples based on the loss value.
\end{abstract}

\begin{IEEEkeywords}
	Batch attention, Image Recognition
\end{IEEEkeywords}

%
\IEEEpeerreviewmaketitle

\section{Introduction}
\begin{figure}[t]
    \begin{center}
        \includegraphics[width=0.5\textwidth]{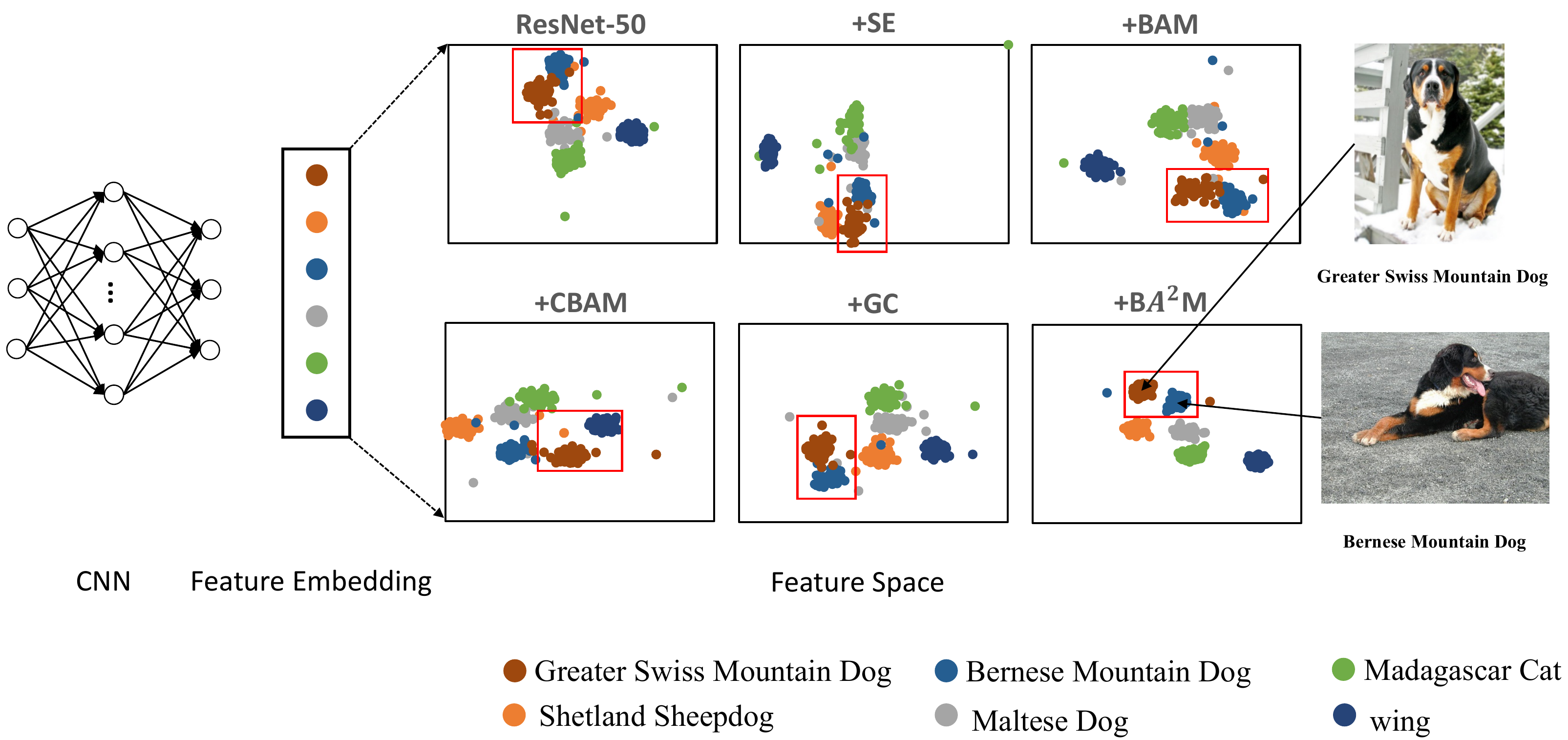}
    \end{center}
    \caption{t-SNE~\cite{t_SNE} visualization of generated features of 700 images refined by different attention mechanisms. The feature clusters refined by ResNet-50+B$\boldsymbol{A}^2$M are more discriminated than ResNet-50~\cite{ResNet}, SE~\cite{SE}, BAM~\cite{BAM}, CBAM~\cite{CBAM}, and GE~\cite{GE}.}
    \label{fig_importance}
\end{figure}

\IEEEPARstart{D}{ue} to the powerful ability of feature enhancement, the attention mechanism has been widely used in designing various network architectures for the image classification task. More specifically, the most popular attention modules could be divided into three types: Channel Attention~\cite{SE,SRM,GE}, Local Spatial Attention~\cite{SK,SGE,BAM,CBAM} and Global Spatial Attention~\cite{GC,AA}. Despite their great successes, these efforts mainly focus on exploring sample-wise attention, whose enhancements are limited within each sample, including the spatial and channel dimensions. However, The optimization of networks is based on batch of images, and all the images contribute equally during the optimization process. When the features of two samples from separated categories are approximately similar, the existing attention mechanism cannot effectively separate them in the feature space. As shown in Fig.~\ref{fig_importance}, we show t-SNE~\cite{t_SNE} visualization of features refined by different attention mechanisms. The images are selected from 14 similar categories in the validation set of ImageNet. When the features of two samples from separated categories are approximately similar, the existing attention mechanism (SE~\cite{SE}, BAM~\cite{BAM}, CBAM~\cite{CBAM}, GC~\cite{GC}) cannot effectively separate them in the feature space and suffers from the overlapping of feature clusters.. For example, the red rectangles in the Fig.~\ref{fig_importance} cover the feature clusters of "Greater Swiss Mountain Dog" and "Bernese Mountain Dog".

\begin{figure*}
    \begin{center}
        \includegraphics[width=0.95\textwidth]{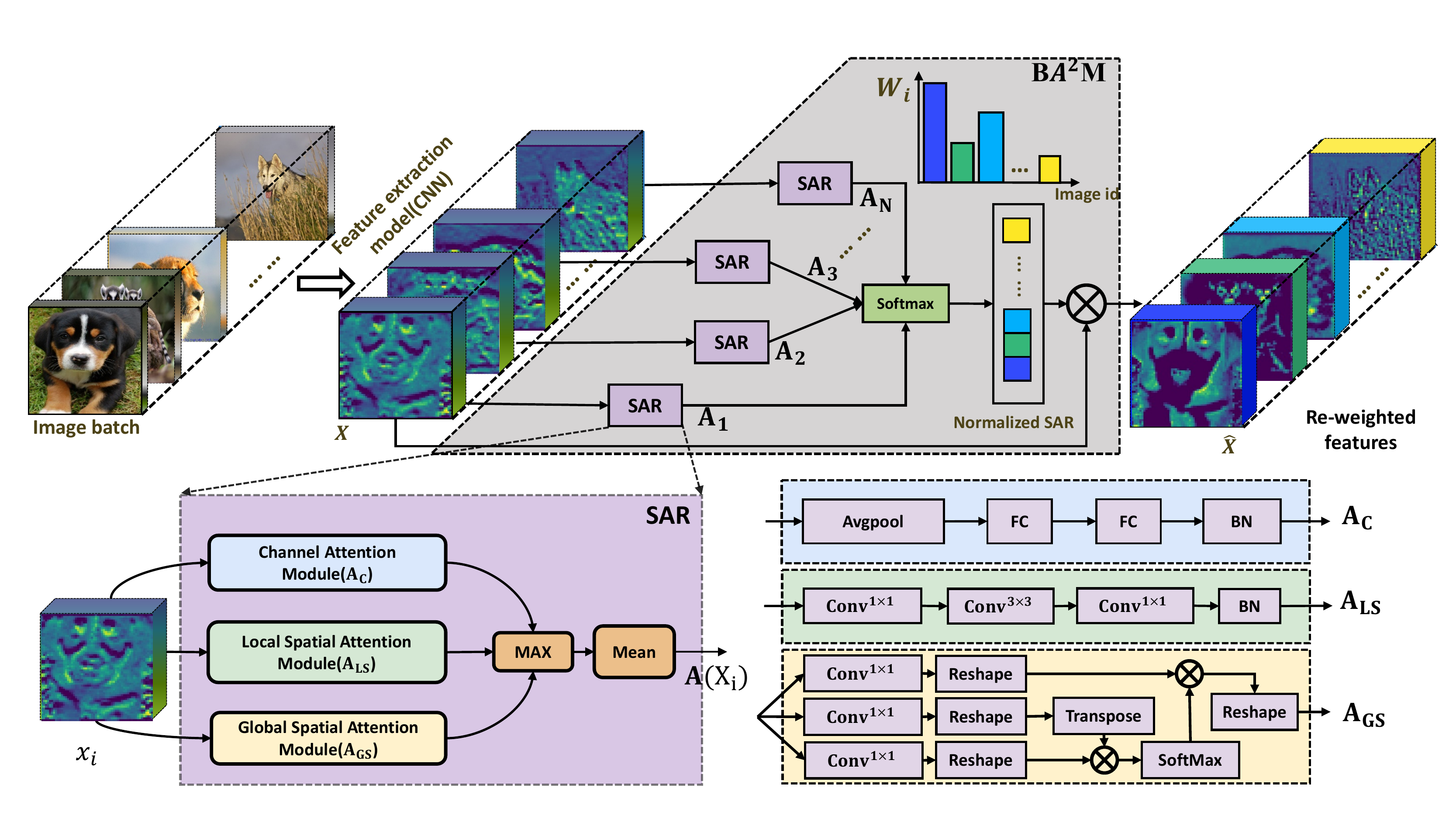}
    \end{center}
    \caption{Illustration of the proposed batch aware attention module (B$\boldsymbol{A}^2$M). The top-row illustrates B$\boldsymbol{A}^2$Net. The bottom-left is a sub-module of B$\boldsymbol{A}^2$M which is used to calculate sample-wise attention representation of each image (SAR). The bottom-right shows the detailed components of three attention modules ($\mathbf{A_{C}}$, $\mathbf{A_{LS}}$, $\mathbf{A_{GS}}$).}
    \label{fig_BSE-Unit}
\end{figure*}

To address the above issues, in this paper, we propose a batch aware attention module (B$\boldsymbol{A}^2$M) to adaptively re-weight the CNN features of different samples within a batch to improve their discrimination in the image classification task.  As shown in Fig.~\ref{fig_importance}, it is seen that the feature clusters of ResNet-50+B$\boldsymbol{A}^2$M are more separated than ResNet-50~\cite{ResNet}, SE~\cite{SE}, BAM~\cite{BAM}, CBAM~\cite{CBAM}, and GE~\cite{GE}, and the feature clusters are more compact. It demonstrates that B$\boldsymbol{A}^2$M could help CNN to obtain more discriminative features. Hence, we first construct a sample-wise attention representation (SAR) for each sample based on three kinds of attention mechanisms including the channel attention (CA), local spatial attention (LSA), and global spatial attention (GSA). The CA captures inter-channel information of the given feature map, LSA generates local information on a limited neighbourhood in the spatial dimension of the given feature map, and GSA explores long-range interactions of pixels of the feature map. Then we feed the SARs of the whole batch to \textit{softmax} layer to obtain normalized weights (\textbf{Normalized-SAR}) which is going to assigned to each sample. The weights serve to increase the features' difference between samples with different categories.

We embed our B$\boldsymbol{A}^2$M into various network architectures and test their performances in the image classification task, including ResNet (ResNet-18/34/50/101~\cite{ResNet,ResNet_v2}, Wide-ResNet-28~\cite{wrn}, ResNeXt-29~\cite{ResNext}), DenseNet-100~\cite{densenet}, ShuffleNetv2~\cite{ShunffleNetv2}, MobileNetv2~\cite{MobileNetv2}, and EfficientsNetB0~\cite{EfficientNet}. Extensive experiments on CIFAR-100 and ImageNet-1K show that B$\boldsymbol{A}^2$M yields significant improvements with negligible complexity increase. It achieves $\textbf{+2.93\%}$ Top-1 accuracy improvement on ImageNet-1K with ResNet-50. In particular,  B$\boldsymbol{A}^2$M significantly outperforms many popular sample-wise attention works such as SE~\cite{SE}, SRM~\cite{SRM}, SK~\cite{SK}, SGE~\cite{SGE}, GE~\cite{GE}, BAM~\cite{BAM}, CBAM~\cite{CBAM}, GC~\cite{GC}, AA~\cite{AA} and SASA~\cite{SASA} in our experiments.

\section{Related work}\label{related_work}
\textbf{Deep feature learning}. Convolutional Neural Networks are potent tools to extract features for computer vision. With the rise of Convolutional Neural Networks (CNN), many sub-fields of computer vision rely on the design of new network structures to be greatly promoted. From VGGNets~\cite{VGG} to Inception models~\cite{Inception}, ResNet~\cite{ResNet}, the powerful feature extraction capabilities of CNN are demonstrated. On the one hand, many works constantly increased the depth~\cite{ResNet_v2}, width~\cite{wrn}, and cardinality~\cite{ResNext} of the network, and reformulate the information flow between network~\cite{densenet}. On the other hand, many works are also concerned with designing new modular components and activation functions, such as depth-wise convolution~\cite{MobileNetv2}, octave convolution~\cite{Octave_Convolution}, ELU~\cite{elu,elu(c)}, and PReLU~\cite{prelu}. These designs can further improve learning and representational properties of CNN.

\textbf{Attention for Feature Enhancement}. In view of the fact that the importance of all extracted features are different, multiple attention mechanisms are developed to enhance the deep feature learning in image recognition task~\cite{Residual_Attention_Net,SE,GE,SRM,SK,SGE,BAM,CBAM,GC,AA}, which focus on highlighting the semantic-aware features within each individual sample. In 2017, Wang \textit{ et al.}~\cite{Residual_Attention_Net} proposed an attention residual module to boost the ability of residual unit to generate attention-aware features. Furthermore, Li \textit{et al.}~\cite{SK} designed a dynamic selection mechanism that allows each neuron pay attention to different size of kernels adaptively. GE~\cite{GE} redistributed the gathered information from long-range feature interactions to local spatial extent. Hu \textit{et al.}~\cite{SE} proposed SE-module to focus on the channel-wise attention by re-calibrating feature maps on channels using signals aggregated from entire feature maps. Lee \textit{et al.}~\cite{SRM} extract statistic information from each channels and re-weight per-channel vial channel-independent information. ~\cite{BAM} and CBAM~\cite{CBAM} refine features beyond the channel with introduced spatial dimensions. Beyond the above attention-works, He \textit{et al.}~\cite{GC} explored the advantages of combination of SE~\cite{SE} and Non-Local~\cite{Non-Local} which further obtains global context information. The above attention mechanisms focus on the intra-features enhancement and ignore the inter-features enhancement.

\section{Approach}\label{approach}
\begin{figure}
    \centering
    \includegraphics[width=0.5\textwidth]{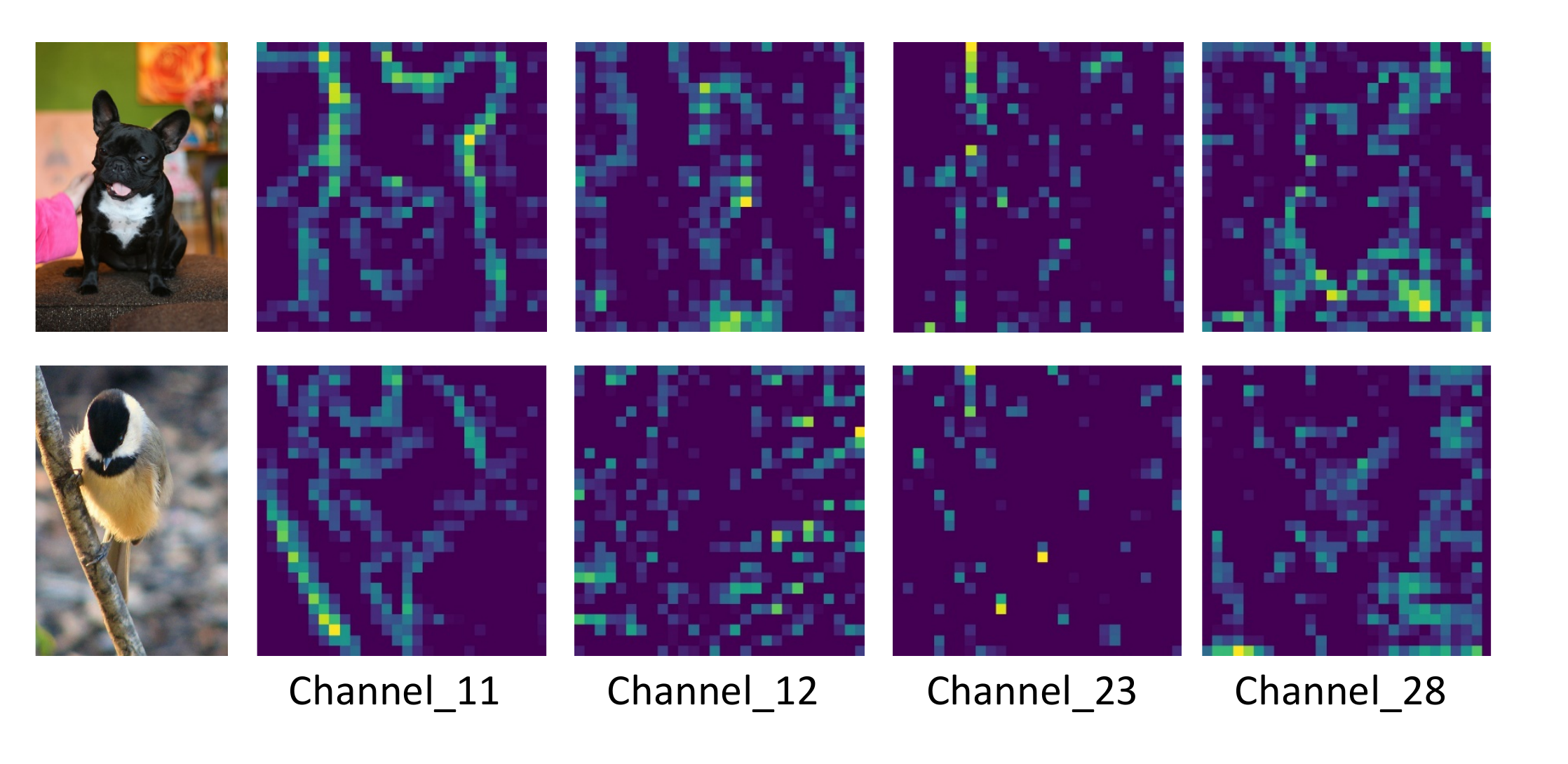}
    \caption{Illustration of some feature maps in the second block of ResNet-50.}
    \label{fig_channel_maps}
\end{figure}

\subsection{Motivation}
Network optimization is carried out with batch mode. In the batch domain with $N$ images, we define the $i_{th}$ input feature $x_{i}$ with the label $y_i$ ($i \in [1,N]$). The original softmax loss for the batch could be described as:
\begin{equation}\label{original_loss_function}
    \begin{aligned}
        L= \frac{1}{N} \sum^{N}_{i} w_{i}\cdot L_{i}
    \end{aligned}
\end{equation}
\begin{equation}\label{original_loss_function_for_single}
    \begin{aligned}
        L_{i}=-\log \left(\frac{e^{f_{y_{i}}}}{\sum^{K}_{j} e^{f_{j}}}\right)
    \end{aligned}
\end{equation}
where $N$ is the number of images in a batch. $w_{i}$ for the original is $1$. $L_{i}$ is the loss for each sample $x_{i}$. where $f_{j}$ denotes the $j_{th}$ elements of the vector of class score $\boldsymbol{f}$. $K$ is the number of classes. $N$ is the number of images in the batch. Equ.~\ref{original_loss_function} means that all the images contribute equally during the optimization process. However, due to the different complexity of the image content, images should have diverse importance while calculating the loss. Focal Loss (FL) \cite{Focal_loss} and online hard example mining (OHEM) \cite{OHEM} show that adaptive adjustment $w_{i}$ can effectively improve the optimization process. OHEM ranks the loss values of each sample in descending order and adds weight to the hard sample with high loss. FL changes the Cross-Entropy loss function and pays more attention with hard ones, so that the samples with large loss values are assigned larger weights, and vice versa. However, these works do not take the image content into account when determining the importance, which acts on the loss directly.

Combine with Equ.~\ref{original_loss_function_for_single}, the Equ.~\ref{original_loss_function} could be modified as follows:
\begin{equation}\label{weighted_loss}
    \begin{aligned}
        L&= \frac{1}{N} \sum^{N}_{i} w_{i}\cdot L_{i} \\&= -\frac{1}{N} \sum^{N}_{i}w_{i}\cdot \log \left(\frac{e^{f_{y_{i}}}}{\sum^{K}_{j} e^{f_{j}}}\right)
        \\&= -\frac{1}{N} \sum^{N}_{i}\log \left( \frac{e^{w_{i}\cdot f_{y_{i}}}} {(\sum^{K}_{j} e^{f_{j}})^{w_{i}}} \right)
    \end{aligned}
\end{equation}
where $f_{y_{i}}=\boldsymbol{W}_{y_{i}}^{T} \cdot x_{i}$. $\boldsymbol{W}$ is the weights of the last fully connected layer. $\boldsymbol{W}_{y_{i}}$ is the $y_{i}$-th column of $\boldsymbol{W}$. The bias $\boldsymbol{b}$ is omit for simplify analysis. $x_{i}$ is the generated feature to the classifier. Scale the loss in Equ.~\ref{weighted_loss}, we could have the following relationship:
\begin{equation}\label{weighted_loss_2}
    \begin{aligned}
        L&= -\frac{1}{N} \sum^{N}_{i}\log \left( \frac{e^{w_{i}\cdot f_{y_{i}}}} {(\sum^{K}_{j} e^{f_{j}})^{w_{i}}} \right) \\&
        \leqslant -\frac{1}{N} \sum^{N}_{i}\log \left( \frac{e^{w_{i}\cdot f_{y_{i}}}} {\sum^{K}_{j} e^{w_{i}\cdot f_{j}}} \right)
        \\&=-\frac{1}{N} \sum^{N}_{i}\log \left( \frac{e^{\boldsymbol{W}_{y_{i}}^{T} (w_{i}\cdot \boldsymbol{x}_{i})}} {\sum^{K}_{j} e^{\boldsymbol{W}_{j}^{T}(w_{i}\cdot \boldsymbol{x}_{i})}} \right)
        \\&=L^{'}
    \end{aligned}
\end{equation}
where inequalities~\ref{weighted_loss_2} is true if $0<w<1$. The proof of inequalities~\ref{weighted_loss_2} is in the \textbf{supplementary material}. $L^{'}$ weights the features directly instead of the loss value of each sample. Besides, $L^{'}$ is the upper bound of original softmax loss $L$. Therefore, if $L^{'}$ goes to 0, then L has to go to 0. That is, the method of feature weighting can get better results than the direct weighting of loss value. The $w_{i}$ is also very important, and it should be related to the content of features. The existing attention mechanisms generate weights based on the content of features. Thus, we try to generate $w_{i}$ based on attention mechanisms.
\subsection{Batch Aware Attention Module (B$\boldsymbol{A}^2$M)}
The attention mechanism is to strengthens information related to optimization goals and suppresses irrelevant information. Existing sample-wise attention are limited to conduct feature enrichment in the spatial or channel dimensions of the feature map. We will extend the attention to batch dimension to generate the weight which is related to the content through the sample-wise attention and collaboratively rectify the features of all samples.
\begin{equation}\label{bwam}
    \begin{aligned}
        w_{i}&=B\boldsymbol{A}^2M(x_{i})
        \\&=Nomalization(\mathbf{A_{C}}(x_{i}), \mathbf{A_{LS}}(x_{i}),\mathbf{A_{GS}}(x_{i}))
    \end{aligned}
\end{equation}
where  $x_{i} \in \mathbb{R}^{C \times H \times W}$ is the feature map. The content of the feature is arranged by three-dimension. Therefore, we decompose the feature from the channel, local spatial and global spatial. $\mathbf{A_{C}}(x_{i})$ denotes the channel attention, $\mathbf{A_{LS}}(x_{i})$ represents the local spatial attention, $\mathbf{A_{GS}}(x_{i})$ expresses the global spatial attention. The three modules are calculated in parallel branches, which are shown in the bottom-left of Fig.~\ref{fig_BSE-Unit}. We will introduce each module in detail in the following sections.

\subsubsection{Channel attention module ($\mathbf{A_{C}}$).}
First, we introduce the channel attention module. The channel contains rich feature information because the feature map $x_{i}$ is arranged by $C$ feature planes. As shown in Fig.~\ref{fig_channel_maps}, each plane has different semantic information~\cite{SE}. In order to exploit the inter-channel information. We follow the practices in SE~\cite{SE} to build a channel attention module. First, we aggregate the feature map of each channel by taking global average pooling (GAP) on the feature map $x_{i}$ and produce a channel vector $GAP(x_{i}) \in \mathbb{R}^{C \times 1 \times 1}$. To estimate attention across channels from channel vector $GAP(x_{i}) $, we use two fully connected layers ($FC_{0}$ and $FC_{1}$). In order to save parameter overhead, we set the output dimension of $FC_{0}$ as  $ C/R$, where $R$ control the overhead of parameters. After $FC_{1}$, we use a batch normalization (BN) layer~\cite{BN} to regulate the output scale. In summary then, we computed the channel attention ($\mathbf{A}_{C} \in \mathbb{R}^{C \times 1 \times 1}$) as:
\begin{equation}\label{channel-branch}
    \begin{aligned}
        \mathbf{A_{C}}(x_i)&= \textbf{BN}(FC_{1}(FC_{0}(GAP(x_i))))\\&=\textbf{BN}(W_1(W_0(GAP(x_i))))
    \end{aligned}
\end{equation}
where $W_0 \in \mathbb{R}^{(C/R) \times C}$,$W_1 \in \mathbb{R}^{C \times (C/R)}$ are trainable parameters and  \textbf{BN} denotes a batch normalization operation.

\subsubsection{Local Spatial attention module ($\mathbf{A_{LS}}$). }
Then, we introduce the local spatial attention module. The spatial dimension usually reflect details of image content. Constructing a local spatial attention module can highlight the area of interest effectively. We cascade one $3 \times 3$ convolution layer between two $1\times 1$ convolution layers to perform local semantic spatial signal aggregation operations. We also utilize a batch normalization layer to adjust the output scale. In short, we compute local spatial attention response ($\mathbf{A_{LS}} \in \mathbb{R}^{C \times H \times W} $) as:
\begin{equation}\label{spatial attention}
    \mathbf{A_{LS}}(x_i) = \textbf{BN}(g^{1 \times 1}_{2}(g^{3\times 3}_{1}(g^{1\times1}_{0}(x_i))))
\end{equation}
where $g$ denotes a convolution operation, and the superscripts denote convolutional filter sizes. We divided $x_{i}$ in channel for $R$ groups for saving memory. Through this spindle-like structural design, we can effectively aggregate local spatial information with small parameter overhead.

\subsubsection{Global Spatial attention module ($\mathbf{A_{GS}}$).}
Finally, we introduce the global spatial attention module. The global spatial emphasizes the long-range relationship between every pixel in the spatial extent. It can be used as a supplement to local spatial attention. Many efforts~\cite{AA,Self-Attention,GC} claim that the long-range interaction can help the feature to be more power. Inspired by the manners of extracting long-range interaction in~\cite{AA}. We simply generate global spatial attention map ($\mathbf{A_{GS}} \in \mathbb{R}^{C \times H \times W} $) as following:
\begin{equation}\label{global_spatial-branch}
    \begin{aligned}
        \mathbf{A_{GS}}(x_i) = \mathit{softmax}(f(x_i)\times (g(x_i))^{T})\times h(x_i)
    \end{aligned}
\end{equation}
where $f$, $g$  and $h$ are convolutional operation with $1 \times 1$ kernel. After reshaping, $f(x_i) \in \mathbb{R}^{WH \times C}$, $g(x_i) \in \mathbb{R}^{WH \times C}$  and $h(x_i)\in \mathbb{R}^{WH \times C}$. The reshaping operations are not shown in Eqn.~\ref{global_spatial-branch} explicitly. However, the specific structure can refer to the bottom-right in Fig.~\ref{fig_BSE-Unit}. $\times$ denotes matrix multiplication. $X$ is processed by $R$ groups for saving memory. Finally, $\mathbf{A_{GS}}(x_i)$  could capture long range interactions to compensate the locality of $\mathbf{A_{LS}}(x_i)$.

\subsubsection{Combination and Excitation for Batch}
In this section, we combine the results generated by the above sample-wised attention modules. First, we normalize the scale of the result spectrum. Then, we choose the result with the strongest activation. Finally, we perform dimension reduction operation on the output of the previous step to one dimension along the channel. In general, $\textbf{A}({x}_{i})$ is formulated as:
\begin{equation}\label{self-importance value}
    \textbf{A}({x}_{i}) = mean(\max(A_{C}(x_{i}), A'_{LS}(x_{i}), A'_{GS}(x_{i})))
\end{equation}
where $A'_{LS}(x_{i})= GAP(A_{LS}(x_{i}))$ and $A'_{GS}(x_{i})= GAP(A_{GS} (x_{i}))$. $A'_{LS}(x_{i})$ and $ A'_{GS}(x_{i}) \in \mathbb{R}^{C \times 1 \times 1}$. $\max$ is the maximum operation, $\textit{mean}$ is the operation of  dimension reduction along channel (\textbf{C}). Thus, the $\textbf{A}({x}_{i}) \in \mathbb{R}^{1 \times 1 \times 1}$. The value $\textbf{A}(x_{i})$ for each $x_{i}$ is related to the complexity of contents.

Based on the above operations, we get sample-wise attention representation (SAR) of each sample, and then we use the SARs of the batch to rectify each sample. The output of batch-wise representation is:
\begin{equation}\label{SAR}
    \begin{aligned}
        \textbf{SAR} =[A_1, A_2, ... A_N] \in \mathbb{R}^{N \times 1 \times 1 \times 1}
    \end{aligned}
\end{equation}
where $N$ represents the number of images in the current batch. To make the  inequalities~\ref{weighted_loss_2} be true, we normalize the vector of SAR in a batch and make SARs more distinguishable. We use \textit{softmax} function, as shown below. After applying \textit{softmax}, each value will be in the interval $(0,1)$. Then, we multiply the SARs with the batch of sample feature maps.
\begin{equation}\label{soft-max}
    \begin{aligned}
        w_{i} = \sigma(\textbf{SAR})_i = \frac{e^{A_i}}{\sum_{j=1}^{N}{e^{A_i}}}
    \end{aligned}
\end{equation}
\begin{equation}\label{SAR_normalized}
    \begin{aligned}
        \textbf{W}=(w_1, w_2, ..., w_N)
    \end{aligned}
\end{equation}
where $\textbf{W}\in \mathbb{R}^{N \times 1 \times 1 \times 1}$ is the vector of weights in current training batch. The batch of refined output $\hat{\textbf{X}}$ could be computed as:
\begin{equation}\label{re-weighted tensors}
    \hat{\textbf{X}} = \textbf{W} \otimes \textbf{X}
\end{equation}
where $\otimes$ denotes element-wise multiplication. It means that each sample $x_{i} \in \mathbb{R}^{C \times H \times W}$ multiplies with a normalized batch-wise attention representation $S_i$.

\subsection{Batch Aware Attention Network (B$\boldsymbol{A}^2$Net)}
\begin{figure}[!t]
    \centering
    \includegraphics[width=0.5\textwidth]{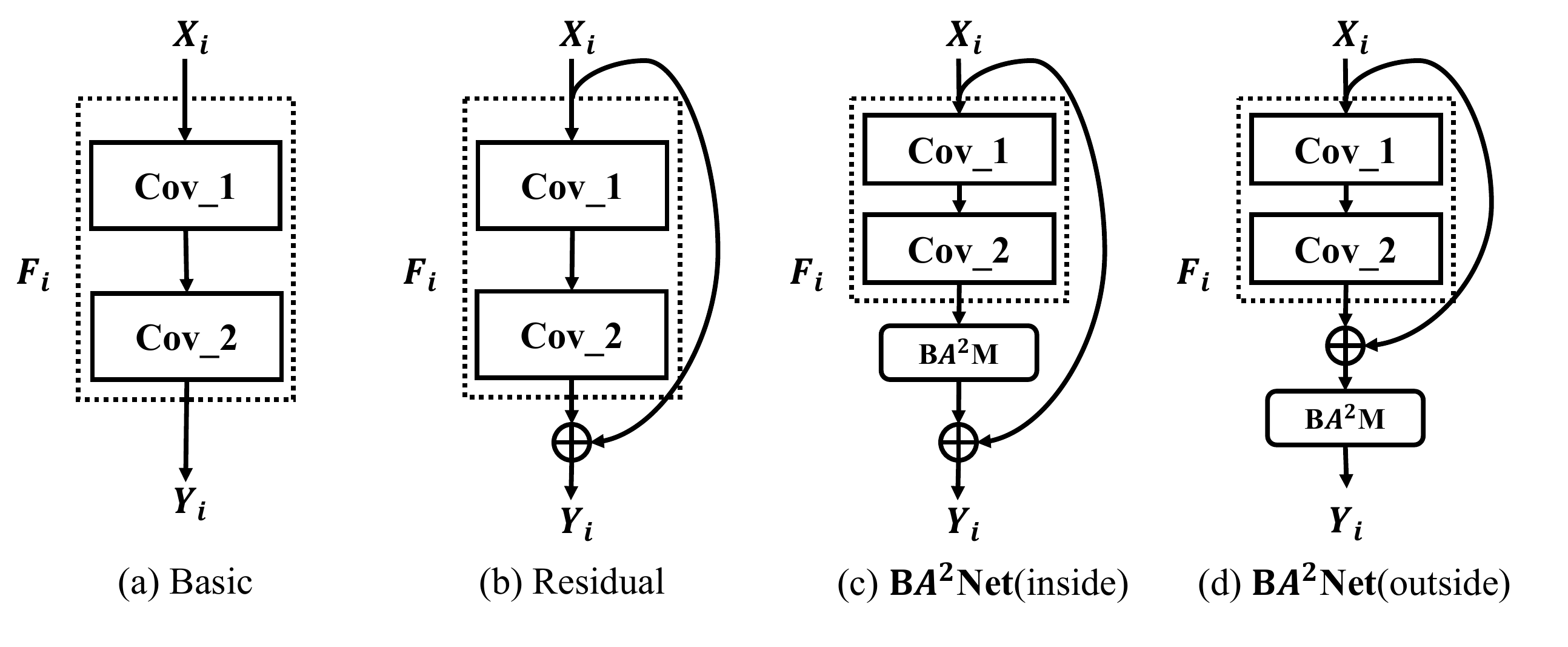}
    \caption{Illustration of the proposed block in B$\boldsymbol{A}^2$Net, basic block and residual block.}
    \label{fig_BSE_Net}
\end{figure}

Given the feature of variations across different layers in a CNN, we try to embed B$\boldsymbol{A}^2$M into different blocks in existing CNNs to build B$\boldsymbol{A}^2$Net. A list of composed blocks could represent most classical ConvNet:
\begin{equation}\label{BSE-Net1}
    \begin{aligned}
        \textit{ConvNet}(X_{i})&=FC  (F_{T}( ... (F_{2}( F_{1}(X_{i})))))
    \end{aligned}
\end{equation}
where $F_{j}$ represents basic or residual block ($j \in [1,T]$, $T$ is the number of blocks) which is shown in Fig.~\ref{fig_BSE_Net} (a) and (b). $X_{i}$ is the $i$-th image in the batch. $FC$ represents the classifier. Activation units ({\textit{e}.\textit{g}.} ReLU \cite{relu}) are omit in Equ.~\ref{BSE-Net1}. We place B$\boldsymbol{A}^2$M between blocks to build B$\boldsymbol{A}^2$Net:
\begin{equation}\label{BSE-Net}
    \begin{aligned}
        H(X_{i})&=FC(w_{Ti}F_{T}( ... (w_{2i}F_{2}(w_{1i}F_{1}(X_{i})))))
    \end{aligned}
\end{equation}
where $H$ represents B$\boldsymbol{A}^2$Net, $F_{j}$ represents the $j_{th}$ residual block, $w_{ji}$ is the $i_{th}$ element of \textbf{W} in the $j_{th}$ block. $T$ is four for ResNet-50.

\subsection{Inference and Complexity analysis}\label{Inference_Complacity}
In this section, we will discuss the inference and complexity of B$\boldsymbol{A}^2$M. B$\boldsymbol{A}^2$M is an adjunct to standard training processes. During the training stage, B$\boldsymbol{A}^2$M equally scales the value of the output vector by weighting samples, thus influencing the loss value. However, during the inference stage, the batch aware weight only affects the absolute value of the output vector, and the rank of elements in the vector does not change. Thus it does not change the final prediction. To verify that B$\boldsymbol{A}^2$M does not affect the results in the inference stage, we activate the B$\boldsymbol{A}^2$M in the ResNet50 on the ImageNet. The size of test batch is $\{1,2,4,8,16,32,64,128,256\}$. All the results are $21.63\%$. Thus, during the inference phase, we test one image a time as the common practice. B$\boldsymbol{A}^2$M generates SAR value for each test image using Eqn.~\ref{self-importance value}. The feature is refined by multiplying with the corresponding SAR. Besides, we deactivate B$\boldsymbol{A}^2$M in the inference phase because we usually infer one image at a time in the common practice where the size of the batch is 1. At this time, excitation B$\boldsymbol{A}^2$M for the batch will lead to over-calculation.

The complexity of B$\boldsymbol{A}^2$M could be divided into three parts, and we use FLOPs and Params to measure them, respectively. We have following equations based on~\cite{time_space}:
\begin{equation}\label{space-complaxity}
    \small
    \begin{aligned}
        \mathbf{Params}\sim O \Bigg (\begin{matrix} \underbrace{ C^{2}\frac{2}{R} } \\ \mathbf{A_{C}}\end{matrix}  + \begin{matrix} \underbrace{ C^{2}(\frac{9+2R}{R^{2}}) } \\ \mathbf{A_{LS}}\end{matrix} + \begin{matrix} \underbrace{  \frac{3C^{2}}{R^{2}}  } \\  \mathbf{A_{GS}}\end{matrix}  \Bigg)
    \end{aligned}
\end{equation}
\begin{equation}\label{time-complaxity}
    \small
    \begin{aligned}
        \textbf{FLOPs} \sim O \Bigg (\begin{matrix} \underbrace{C^{2} \frac{2}{R} } \\  \mathbf{A_{C}}\end{matrix}  + \begin{matrix} \underbrace{ HWC^{2}(\frac{9+2R}{R^{2}}) } \\  \mathbf{A_{LS}}\end{matrix} +\\
        \begin{matrix} \underbrace{ HWC^{2}(\frac{3}{R^{2}}) + 2(HW)^{2}(\frac{(2C-R)}{R})} \\  \mathbf{A_{GS}}\end{matrix}  \Bigg)
    \end{aligned}
\end{equation}
where the input feature map is $\in \mathbb{R}^{N \times C \times H \times W}$ . We can see that based on Eqn.~\ref{time-complaxity} and \ref{space-complaxity}, $\textbf{R}$ plays an important role in controlling the complexity of the B$\boldsymbol{A}^2$M. We will discuss the influence of $\textbf{R}$ in ablation experiments. The Flops of $ \mathbf{A_{GS}}$ in Eqn.~\ref{time-complaxity} have two parts. The first part is the time-complexity of three convolutional operations, and the second part is related to two matrix multiplication operations. However, there are no extra parameters in matrix multiplication, Thus the parameter measure in $ \mathbf{A_{GS}}$ is $\frac{3C^{2}}{R^{2}} $.

\section{Experiments}\label{experiment}
In order to evaluate the effect of proposed B$\boldsymbol{A}^2$M, we conduct several experiments on  two widely-used image classification datasets:  CIFAR-100~\cite{Cifar100} and  ImageNet-1k~\cite{ImageNet}. PyTorch-0.11.0 library~\cite{pytorch} is utilized to implement all experiments on NVIDIA TITANX GPU graphics cards. CIFAR-related and ImageNet-related experiments are on a single card and eight cards, respectively. For each configuration, we repeat the experiments for \textbf{five} times with different random seeds and report the median result and the best (per run) validation performance of the architectures over time. The setting of hyper-parameters is the same as~\cite{BAM}.

\subsection{Experiments on CIFAR-100}
CIFAR-100~\cite{Cifar100} is a tiny natural image dataset, which contains 100 different classes. Each image is in size of $32 \times 32$. There are 500 images for training and 100 images validation per class. We adopt some simple data augmentation strategies, such as random crops, random horizontal flips, and mean-variance normalization.We performed image classification with a range of ResNet architecture and its variants: ResNet-50~\cite{ResNet}, ResNet-101~\cite{ResNet}, PreResNet-110~\cite{Preresnet}, Wide-ResNet-28 (WRN-28)~\cite{wrn} and ResNeXt-29~\cite{ResNext}. We reported classification error on validation set as well as model size (Params) and computational complexity (Flops).

Results are shown in Table.~\ref{tab:Cifar100-results-different-networks}. It could be concluded from the results that B$\boldsymbol{A}^2$M could consistently improve the classification performance of the network regardless of the network structure without increasing the amount of calculation. From the above results, we can conclude that B$\boldsymbol{A}^2$M is an effective method.

\begin{table}
    \centering
    \resizebox{0.45\textwidth}{!}{%
        \begin{tabular}{lccc}
            \toprule[1.0pt]
            Architecture & Params(M) & FLOPs(G) &  Error (\%) \\ \midrule
            ResNet-50~\cite{ResNet} & 23.71 & 1.22 & 21.49 \\
            +B$\boldsymbol{A}^2$M (ours) &24.21  & 1.33  &  \textbf{17.60\scriptsize{(-3.89)}}

            \\ \midrule

            ResNet-101~\cite{ResNet} & 42.70 & 2.44 & 20.00 \\
            {{+B$\boldsymbol{A}^2$M} (ours)} & 42.32 & 2.55 &  \textbf{17.88\scriptsize{(-2.12)}}

            \\ \midrule

            {PreResNet-110}~\cite{Preresnet} & 1.73 & 0.25 & 22.22 \\
            {{+B$\boldsymbol{A}^2$M} (ours)} & 1.86 & 0.25 & \textbf{21.41\scriptsize{(-0.81)}}

            \\ \midrule

            {WRN-28(w=8)}~\cite{wrn} & 23.40 & 3.36 & 20.40 \\
            {{+B$\boldsymbol{A}^2$M} (ours)} & 23.44 & 3.46 &\textbf{17.95\scriptsize{(-2.45)}}

            \\ \midrule

            {WRN-28(w=10)}~\cite{wrn} & 36.54 & 5.24 & 18.89 \\
            {{+B$\boldsymbol{A}^2$M} (ours)} & 36.59 & 5.39 &  \textbf{17.30\scriptsize{(-1.59)}}

            \\ \midrule

            {ResNeXt-29}~\cite{ResNext} & 34.52 & 4.99 & 18.18 \\
            {{+B$\boldsymbol{A}^2$M} (ours)} & 34.78 & 5.53 & \textbf{16.10\scriptsize{(-2.08)}}

            \\ \bottomrule[1pt]
        \end{tabular}%
    }
    \caption{Image classification results on CIFAR-100 ~\cite{Cifar100} across a range of ResNet architectures and its variants. The results of compared methods could be find in corresponding papers.}
    \label{tab:Cifar100-results-different-networks}
\end{table}

\subsection{Experiments on ImageNet-1K}\label{imagenet2015}
In order to further validate the effectiveness of B$\boldsymbol{A}^2$M, in this section, we perform image classification experiments on more challenging 1000-class ImageNet dataset~\cite{ImageNet}, which contains about 1.3 million training color images as well as 50,000 validation images. We use the same data augmentation strategy as~\cite{ResNet, ResNet_v2} for training and a single-crop evaluation with the size of $224\times 224$ in the testing phase. We report FLOPs and Params for each model, as well as the top-1 and top-5 classification errors on the validation set. We use a range of ResNet architectures and their variants: ResNet-18~\cite{ResNet}, ResNet-34~\cite{ResNet}, Wide-ResNet-18 (WRN-18)~\cite{wrn}, ResNeXt-50~\cite{ResNext} and ResNeXt-101~\cite{ResNext}.

The results are shown in Table.~\ref{ImageNet-results-different-networks}. The networks with B$\boldsymbol{A}^2$M outperform the baseline network in performance, which is an excellent proof that B$\boldsymbol{A}^2$M could improve the discrimination of features in such a challenging dataset. It is worth noting that it could be negligible in the overhead of both parameters and computation, which shows that B$\boldsymbol{A}^2$M could significantly improve the model capacity efficiently. When B$\boldsymbol{A}^2$M is combined with WRN, the increment of the parameter is a little bigger. The abnormal increment is mainly due to the wide of $WRN$ significantly increases the number of channel $C$ in the convolutional layer.

\begin{table}
    \centering
    \resizebox{0.45\textwidth}{!}{%
        \begin{tabular}{lcccc}
            \toprule[1.0pt]
            {Architecture} & Params.(M) & FLOPs(G) & \textbf{Top1-Error(\%)}  \\ \midrule
            {ResNet-18~\cite{ResNet}} & 11.69 & 1.81 & 29.60 \\
            {{+B$\boldsymbol{A}^2$M} (ours)} & 12.28 & 1.92 & \textbf{ 28.62\scriptsize{(-0.98)} }
            \\ \midrule[1.0pt]

            {ResNet-34~\cite{ResNet}} & 21.80 & 3.66 & 26.69 \\
            {{+B$\boldsymbol{A}^2$M} (ours)} & 21.92 & 3.77 &  \textbf{25.15\scriptsize{(-1.54)}}
            \\ \midrule[1.0pt]

            {WRN-18(w=1.5)~\cite{wrn}} & 25.88 & 3.87 & 26.85 \\
            {{+B$\boldsymbol{A}^2$M} (ours)} & 28.43& 4.69 &  \textbf{23.60\scriptsize{(-3.25)}}
            \\ \midrule[1.0pt]

            {WRN-18(w=2)~\cite{wrn}} & 45.62 & 6.70 & 25.63 \\
            {{+B$\boldsymbol{A}^2$M} (ours)} & 47.88 & 7.56 & \textbf{23.60\scriptsize{(-2.03)}}
            \\ \midrule[1.0pt]

            {ResNeXt-50~\cite{ResNext}} & 25.03 & 3.77 & 22.85 \\
            {{+B$\boldsymbol{A}^2$M} (ours)} & 27.25& 4.47 &  \textbf{20.98\scriptsize{(-1.87)}}
            \\ \midrule[1.0pt]

            {ResNeXt-101~\cite{ResNext}} & 44.18 & 7.51 & 21.54 \\
            {{+B$\boldsymbol{A}^2$M} (ours)} & 45.51 & 8.23 &\textbf{ 20.05\scriptsize{(-1.49)}}
            \\ \bottomrule[1.0pt]
        \end{tabular}%
    }
    \caption{Image classification results on ImageNet-1K dataset~\cite{ImageNet} across a range of ResNet architectures and its variants.}
    \label{ImageNet-results-different-networks}
\end{table}

\subsection{Comparison with Focal Loss and OHEM}
In this section, we compare B$\boldsymbol{A}^2$M with Focal Loss \cite{Focal_loss} and online hard example mining (OHEM) \cite{OHEM}. B$\boldsymbol{A}^2$M re-weights samples in a batch by calculating the attention value of the samples, this could also solve the problem of class imbalance in the object detection field to some extent. Focal loss and OHEM are specifically designed to solve the problem of class imbalance. OHEM selects a candidate ROI with a massive loss to solve the category imbalance problem. Focal loss achieves the effect of sample selection by reducing the weight of the easily categorized samples so that the model more focus on difficult samples during training. We perform experiments on the union set of  PASCAL VOC 2007 trainval and PASCAL VOC 2012 trainval (VOC0712) and evaluate on the PASCAL VOC 2007 test set.

For Focal Loss, We adopt RetinaNet \cite{Focal_loss} as our detection method and ImageNet pre-trained ResNet-50 as our baseline networks. Then we replace Focal Loss with SoftmaxLoss and change the ResNet-50 to ResNet-50 with B$\boldsymbol{A}^2$M. Results are shown in Table \ref{tab:VOC_Focal_loss},  the Focal loss is lower than B$\boldsymbol{A}^2$M on both Recall and mAP.

\begin{table}
    \centering
    \resizebox{0.45\textwidth}{!}{%
        \begin{tabular}{llcc}
            \toprule
            Detector & Method & Recall (\%)  & mAP (\%)  \\
            \midrule
            RetinaNet & Focal Loss & 96.27 & 79.10 \\
            RetinaNet & B$\boldsymbol{A}^2$M(ours) &  \textbf{97.12\scriptsize{(+0.85)}} & \textbf{79.60\scriptsize{(+0.50)}} \\
            \bottomrule
        \end{tabular}
    }
    \caption{\textbf{Object detection} results compared with Focal Loss. Baseline detector is RetinaNet. Backbone is ResNet-50 pre-trained on ImageNet.}
    \label{tab:VOC_Focal_loss}
\end{table}

For OHEM, We adopt Faster-RCNN \cite{faster_rcnn} as our detection method and ImageNet pre-trained ResNet-50 as our baseline network. From the results summarized in Table \ref{tab:VOC0712_BSE_OHEM}. We can see that the Recall and mAP of B$\boldsymbol{A}^2$M are both optimal. Compared with the traditional methods of re-weight samples based on the loss value, the weights generated by B$\boldsymbol{A}^2$M are better.

\begin{table}
    \centering
    \resizebox{0.45\textwidth}{!}{%
        \begin{tabular}{llcc}
            \toprule
            Detector &Method&  Recall (\%) & mAP (\%) \\
            \midrule
            Faster R-CNN&-&92.97 & 80.10\\
            Faster R-CNN&OHEM \cite{OHEM}& 90.57 & 80.50\\
            Faster R-CNN&B$\boldsymbol{A}^2$M (ours)& \textbf{93.99\scriptsize{(+1.02)}} & \textbf{81.00\scriptsize{(+0.90)}}\\
            \bottomrule
        \end{tabular}
    }
    \caption{\textbf{Object detection} results compared with OHEM. Baseline detector is Faster R-CNN. Backbone is ResNet-50 pre-trained on ImageNet.}
    \label{tab:VOC0712_BSE_OHEM}
\end{table}

\subsection{Ablation Experiments}\label{ablation_experiment}

\subsubsection{Comparison with Sample-wise Attention}

In this section, we systematically compare B$\boldsymbol{A}^2$M with some attention works through image classification on ImageNet-1K~\cite{ImageNet}. We choose SE~\cite{SE} and SRM~\cite{SRM}, SK~\cite{SK}, SGE~\cite{SGE} and GE~\cite{GE}), BAM~\cite{BAM}, CBAM~\cite{CBAM}, GC~\cite{GC}, AA~\cite{AA} and SASA~\cite{SASA} as compared methods. These works mainly focus on exploring sample-wise attention, whose enhancements are limited within each individual sample including the spatial and channel dimensions. We choose ResNet-50 and ResNet-101 as baseline networks and replace attention modules at each block. All models maintain the same parameter settings during training.

\begin{table*}
    \centering
    \resizebox{0.95\textwidth}{!}{%
        \begin{tabular}{lccccccccc}
            \toprule
            \multirow{2}{*}{} & \multicolumn{4}{c}{\textbf{ResNet-50}} & \multicolumn{4}{|c}{\textbf{ResNet-101}} \\ \cmidrule(l){2-9}
            & {Params.(M)} & {FLOPs(G)} & {Top1-Error(\%)} & {Top5-Error(\%)} & \multicolumn{1}{|c}{{Params.(M)}} & \multicolumn{1}{c}{{FLOPs(G)}} & \multicolumn{1}{c}{{Top1-Error(\%)}} & \multicolumn{1}{l}{{Top5-Error(\%)}} \\ \midrule
            {Baseline} & 25.56 & 3.86 & 24.56 & 7.50 & 44.55 & 7.57 & 23.38 & 6.88 \\ \midrule

            {+SE}~\cite{SE} & 28.09 & 3.86 & 23.14 & 6.70 & 49.33 & 7.58 & 22.35 & 6.19 \\
            {+SRM}~\cite{SRM} & 25.62 & 3.88 & 22.87 & 6.49 & 44.68 & 7.62 & 21.53 & 5.80 \\ \midrule

            {+SK}~\cite{SK} & 26.15 & 4.19 & 22.46 & 6.30 & 45.68 & 7.98 & 21.21 & 5.73 \\
            {+SGE}~\cite{SGE} & 25.56 & 4.13 & 22.42 & 6.34 & 44.55 & 7.86 & 21.20 & 5.63\\
            {+GE}~\cite{GE} & 31.20 & 3.87 & 22.00 & 5.87 & 33.70 & 3.87 & 21.88 & 5.80 \\ \midrule

            {+BAM}~\cite{BAM} & 25.92 & 3.94 & 24.02 & 7.18 & 44.91 & 7.65 & 22.44 & 6.29 \\
            {+CBAM}~\cite{CBAM} & 28.09 & 3.86 & 22.66 & 6.31 & 49.33 & 7.58 & 21.51 & 5.69 \\ \midrule

            {+GC}~\cite{GC} & 28.08 & 3.87 & 22.30 & 6.34 & 49.36 & 7.86 & 25.36 & 7.93 \\
            {+AA}~\cite{AA} & 25.80 & 4.15 & 22.30 & 6.20 & 45.40 & 8.05 & 21.30 & 5.60 \\ \midrule

            {+SASA}~\cite{SASA} & 18.00 & 7.20 & 22.40 & - & - & - & - & - \\ \midrule
            {{+B$\boldsymbol{A}^2$M} (ours)}& 26.21\scriptsize{(+0.65)} & 4.32\scriptsize{(+0.46)} & \textbf{21.63\scriptsize{(-2.93)}} & \textbf{5.80\scriptsize{(-1.70)}} & 45.87\scriptsize{(+1.32)} & 8.05\scriptsize{(+0.48)} & \textbf{20.85\scriptsize{(-2.53)}} & \textbf{5.58\scriptsize{(-1.30)}} \\ \bottomrule
        \end{tabular}%
    }
    \caption{\textbf{Image classification} results on ImageNet-1K dataset ~\cite{ImageNet} with ResNet-50 and ResNet-101 across a range of attention mechanisms:SE~\cite{SE}, SRM~\cite{SRM}, GE~\cite{GE}, SK~\cite{SK}, SGE~\cite{SGE}, BAM~\cite{BAM}, CBAM~\cite{CBAM}, GC~\cite{GC}, AA~\cite{AA} and SASA~\cite{SASA}.}
    \label{ImageNet-results-different-attention}
\end{table*}

The results are shown in Table.~\ref{ImageNet-results-different-attention}, It could be inferred from the results that B$\boldsymbol{A}^2$M could significantly improve the performance of baseline and B$\boldsymbol{A}^2$M is superior to sample-wise attention mechanisms in improving the performance of baseline with little overhead on parameters and computation. Besides, In Table.~\ref{ImageNet-results-different-networks} and Table.~\ref{ImageNet-results-different-attention}, we have performed experiments with ResNet18/34/50/101 on ImageNet. The gains are significant improved from 0.98\% for ResNet18 to 2.53\% for ResNet101 as the network architecture becoming deeper.

\subsubsection{Cooperation with Sample-wise Attention}

In this section, we perform experiments with a combination of batch-with and other sample-wise attention methods. We assume that the combination with other sample-with attention methods should improve model performance even more.  We choose SE~\cite{SE}, CBAM~\cite{CBAM}, BAM~\cite{BAM} to be the instance methods. We use ResNet-50 as the baseline networks. The results are shown in Table.~\ref{tab:Cifar100-results-with-other attentions}. The combination of B$\boldsymbol{A}^2$M with other attention methods could improve performance even more. Especially, the combination of BAM and B$\boldsymbol{A}^2$M even gets the classification error of $17.00\%$.

\begin{table}
    \centering
    \resizebox{0.45\textwidth}{!}{%
        \begin{tabular}{lccc}
            \toprule[1.0pt]
            Architecture & Params.(M) & FLOPs(G) &  Error (\%) \\ \midrule
            ResNet-50~\cite{ResNet} & 23.71 & 1.22 & 21.49 \\
            +SE~\cite{SE} & 26.24 & 1.23 & 20.72 \\
            +CBAM~\cite{CBAM} & 27.44 & 1.34 & 21.01 \\
            +BAM~\cite{BAM} & 24.07 & 1.25 & 20.00 \\
            +B$\boldsymbol{A}^2$M  &24.21  & 1.33  & 17.60 \\
            +SE+B$\boldsymbol{A}^2$M & 26.74 & 1.34 & 17.52 \\
            +CBAM+B$\boldsymbol{A}^2$M & 27.94 & 1.45 & 17.43 \\
            +BAM+B$\boldsymbol{A}^2$M  & 24.57 & 1.36 & \textbf{17.00} \\
            \bottomrule[1pt]
        \end{tabular}%
    }
    \caption{Image classification results with other attention modules on CIFAR-100~\cite{Cifar100}. }
    \label{tab:Cifar100-results-with-other attentions}
\end{table}

\subsubsection{The position of B$\boldsymbol{A}^2$M}
The location for B$\boldsymbol{A}^2$M in the residual-like network has two choices. The first is to embed B$\boldsymbol{A}^2$M inside the block, as shown in Fig.~\ref{fig_BSE_Net} (c ). The other is to embed B$\boldsymbol{A}^2$M between blocks, as shown in Fig.~\ref{fig_BSE_Net} (d). In this section, we conduct experiments to determine the embed mode. The results are exhibited in Table.~\ref{tab:bse_position}. The average error and standard deviation of \textbf{five} random runs are recorded and the best results are in \textbf{bold}. We found that placing B$\boldsymbol{A}^2$M between blocks is better than inside blocks in improving the performance (17.60\% Vs 17.71\%) and stability of the model (0.15 Vs 0.40). Thus,  we assign B$\boldsymbol{A}^2$M between blocks to build different B$\boldsymbol{A}^2$Net.

\begin{table}
    \centering
    \begin{tabular}{lcc}
        \toprule
        Position    &  Error(\%)  & std    \\ \midrule
        Inside (Fig.~\ref{fig_BSE_Net}(c))  & 17.71   & 0.40         \\
        Between (Fig.~\ref{fig_BSE_Net}(d)) & \textbf{17.60} &  \textbf{0.15} \\ \bottomrule
    \end{tabular}%
    \caption{The results of the position of B$\boldsymbol{A}^2$M. }
    \label{tab:bse_position}
\end{table}

\subsubsection{Sensitivity of Hyper-Parameters}

In this section, we empirically show the effectiveness of design choice. For this ablation study, we use ResNet-50 as the baseline architecture and train it on the CIFAR-100 dataset.

\begin{figure}
    \centering
    \includegraphics[width=0.5\textwidth]{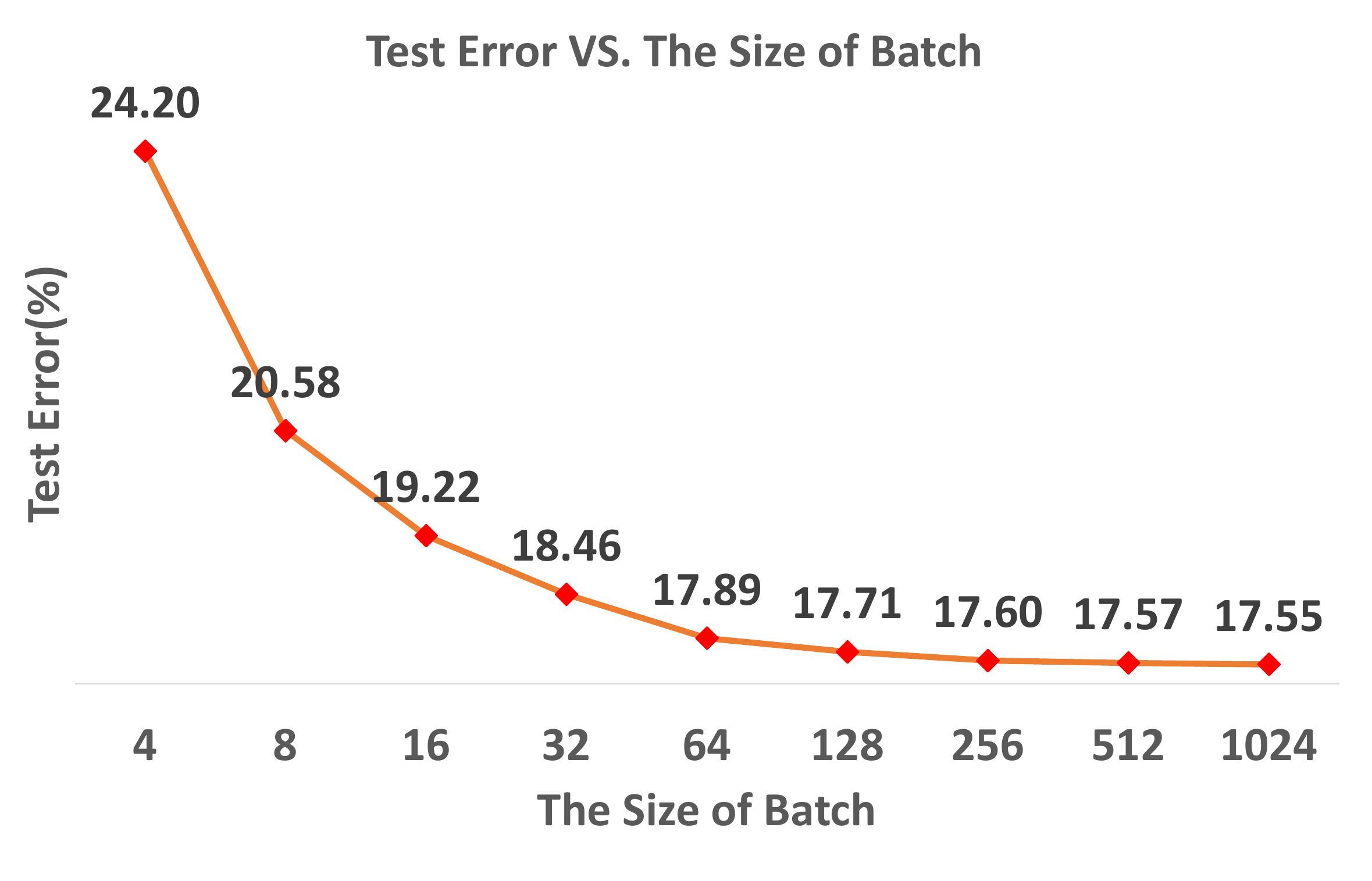}
    \caption{Test error vs The size of Batch. All models are trained on one GPU and learning rate changed linearly.}
    \label{fig_test_error_vs_batchsize}
\end{figure}

\textbf{The size of Batch (N) }. We perform experiments with $N \in \{4,8,16,32,64,128,256\}$ under the memory limitation of GPU (12G). The learning rate (lr) for $N=128$ is $0.1$. For other $N$, the lr changes linearly. Each setting runs for \textbf{five} times, and we reported mean test error. As shown in Fig.~\ref{fig_test_error_vs_batchsize}, classification performance continuously improves with larger \textbf{N}, which indicates that B$\boldsymbol{A}^2$M is more efficient when more samples are involved in re-weighting. In experiments, we set \textbf{N} to be 256 to balance computing efficiency and hardware overhead.

\textbf{Reduction (R)}.We conduct experiments to determine hyper-parameters in B$\boldsymbol{A}^2$M, which is the reduction ratio (\textbf{R}). \textbf{R} is used to control the number of channels in three modules, which enable us to control the capacity and the overhead of B$\boldsymbol{A}^2$M. The minimum number of channels in the ablation experiment is 32. In Table.~\ref{tab:Ablation_reduction}, we compare mean performance for \textbf{five} random runs of $R \in \{2, 4, 8, 16, 32\}$ and corresponding model size (Params) and computational complexity (Flops, Madds). Interestingly, as \textbf{R} increases, the overhead of computation and model size continue to decrease. However, the corresponding performance first drops and then rises. When $R$ is 32, the corresponding performance is the best. To balance model complexity and performance, we choose \textbf{R} to be 32.

\begin{table}
    \centering
    \resizebox{0.45\textwidth}{!}{%
        \begin{tabular}{ccccc}
            \toprule
            \textbf{R} & \textbf{Params(MB)} & \textbf{Flops(GB)} & \textbf{ Error(\%)}  & \textbf{std} \\ \midrule
            \textbf{2} & 70.80     & 2.86      & 17.71 & 0.40      \\
            \textbf{4} & 40.65   & 1.78      & 18.04   & 0.19        \\
            \textbf{8} & 30.12   & 1.47     & 18.13   & 0.10        \\
            \textbf{16}& 25.99   & 1.37     & 18.01   & 0.13        \\
            \textbf{32}& \textbf{24.21}     & \textbf{1.33}     & \textbf{17.60}  & 0.15        \\ \bottomrule
        \end{tabular}%
    }
    \caption{Results with different reduction in $\{2,4,8, 16,32\}$. We train ResNet-50 as the base network on CIFAR-100.}
    \label{tab:Ablation_reduction}
\end{table}

\textbf{Design Choices of B$\boldsymbol{A}^2$M}. We perform experiments to determine the design choices of B$\boldsymbol{A}^2$M. There are three attention modules in B$\boldsymbol{A}^2$M. Thus, we have seven combinations of attention mechanisms. The results are shown in Table.~\ref{tab:Combination_within_BSE_Unit}. From the second row to the fourth row, we can find that \textit{LSA} ( $17.86\%$ ) plays the most crucial role in B$\boldsymbol{A}^2$M. Besides, the combination of LSA and GSA gets $17.78\%$ performance which is better than single LSA. The above results demonstrate that the long-range interactions could compensate the local spatial attention. Furthermore, the full version of B$\boldsymbol{A}^2$M achieves the best performance.

\begin{table}
    \centering
    \resizebox{0.4\textwidth}{!}{%
        \begin{tabular}{ccccc}
            \toprule
            {CA} & {LSA} & {GSA} & { Error (\%)} & {Std} \\ \midrule
            \checkmark &  &  & 18.06 & 0.20 \\
            & \checkmark &  & 17.86 & 0.30 \\
            &  & \checkmark & 18.18 & 0.29 \\
            \checkmark & \checkmark &  & 17.71 & 0.30 \\
            \checkmark &  & \checkmark & 18.00 & 0.23 \\
            & \checkmark & \checkmark & 17.78 & 0.28 \\
            \checkmark & \checkmark & \checkmark &  \textbf{17.60} & \textbf{0.15} \\
            \bottomrule
        \end{tabular}
    }
    \caption{The results of different design choices of B$\boldsymbol{A}^2$M. The last row is the result of B$\boldsymbol{A}^2$M.}
    \label{tab:Combination_within_BSE_Unit}
\end{table}

\section{Conclusion}\label{conclusion}
We have presented a batch-wise attention module(B$\boldsymbol{A}^2$M). It provides new insight into how the attention mechanism can enhance the discrimination power of features. B$\boldsymbol{A}^2$M rectifies the features by sample-wise attention across a batch. It is a lightweight module and can be embed between blocks of a CNN. To verify B$\boldsymbol{A}^2$M's efficacy, we conducted extensive experiments with various state-of-the-art models. The results proofed that B$\boldsymbol{A}^2$M can boost the performance of the baseline and outperform other methods.

\appendices
\subsection{Supplementary material}
    In this supplementary material, we give the proof of inequalities.4 and present more results and conduct some object detection and instance segmentation experiments to verify the effectiveness of B$\boldsymbol{A}^2$M.

\subsection{Proof of Equation}
\textbf{Lemma 1.} For variable $x$ and $y$ hold
\begin{equation}\label{Nis2_1}
    \begin{aligned}
        (x+y)^{w} < x^{w}+y^{w}
    \end{aligned}
\end{equation}
where
\begin{equation}\label{Nis2_2}
    \begin{aligned}
        0<w<1, x,y>0
    \end{aligned}
\end{equation}
\textit{proof.}  Let's define a function $f(t)$ as follow:
\begin{equation}\label{Nis2_3}
    \begin{aligned}
        f(t)=(t+1)^w-t^{w}-1
    \end{aligned}
\end{equation}
where
\begin{equation}\label{Nis2_4}
    \begin{aligned}
        t\geqslant 0, 0<w<1
    \end{aligned}
\end{equation}
The derivative of $f(t)$ is:
\begin{equation}\label{Nis2_5}
    \begin{aligned}
        f^{'}(t)&=w(t+1)^{w-1}-wt^{w-1}\\&= w[(t+1)^{w-1}-t^{w-1}]\\&<0
    \end{aligned}
\end{equation}
The last line with inequality sign of Equ.~\ref{Nis2_5} is because the function $t^{(w-1)}$ is monotonic decreasing function when $-1<w-1<0$.

Therefore, $f(t)$ is also a monotonic decreasing function. For each element $t>0$, we have the following relation:
\begin{equation}\label{Nis2_6}
    \begin{aligned}
        f(t) &=(t+1)^w-t^{w}-1 \\& \leq f(0)=0
    \end{aligned}
\end{equation}
Therefore,
\begin{equation}\label{Nis2_7}
    \begin{aligned}
        (t+1)^w < t^{w}+1
    \end{aligned}
\end{equation}
where
\begin{equation}\label{Nis2_8}
    \begin{aligned}
        t> 0, 0<w<1
    \end{aligned}
\end{equation}
Let $t=x/y$, where $x>0$, $y>0$. We will get the following conclusion:
\begin{equation}\label{Nis2_9}
    \begin{aligned}
        (\frac{x}{y}+1)^w < (\frac{x}{y})^{w}+1
    \end{aligned}
\end{equation}
Due to $y>0$ and $0<w <1$, thus, $y^{w}>0$. We use $y^{w}$ multiply the both sides of Equ.~\ref{Nis2_9}. Therefore, we get
\begin{equation}\label{Nis2_10}
    \begin{aligned}
        (x+y)^{w}< x^{w}+y^{w}
    \end{aligned}
\end{equation}

\textbf{Lemma 2.} For variable $x_{i}$ hold
\begin{equation}\label{Nis3_1}
    \begin{aligned}
        (\sum ^{N}_{i}x_{i})^{w}\leqslant \sum ^{N}_{i}(x_{i})^{w}
    \end{aligned}
\end{equation}
where
\begin{equation}\label{Nis3_2}
    \begin{aligned}
        0<w<1, x_{i}>0, N \in \mathbb{N}
    \end{aligned}
\end{equation}
\textit{proof.} Let's employ mathematical induction with the number $N$.
\begin{description}
    \item[STEP1], When $N=1$, The Equ.~\ref{Nis3_1} gets the equal sign.

    \item[STEP2], When $N=2$, The Equ.~\ref{Nis3_1} could be hold by \textbf{Lemma 1}.

    \item[STEP3], if $N=K$($K\in \mathbb{N}$ and $K>2$), The Equ.~\ref{Nis3_1} is true. We get
    \begin{equation}\label{Nis3_3}
        \begin{aligned}
            (\sum ^{K}_{i}x_{i})^{w}\leqslant \sum ^{K}_{i}(x_{i})^{w}
        \end{aligned}
    \end{equation}
    When $N=K+1$,
    \begin{equation}\label{Nis3_4}
        \begin{aligned}
            (\sum ^{K+1}_{i}x_{i})^{w}&=((\sum ^{K}_{i}x_{i})+x_{(K+1)})^{w}\\& \leqslant \sum ^{K}_{i}(x_{i})^{w}+x^{w}_{(K+1)},\\&=\sum ^{K+1}_{i}(x_{i})^{w}
        \end{aligned}
    \end{equation}
    Therefore, Equ.~\ref{Nis3_1} holds when $N=K+1$.

    \item[CONCLUTION],

    The Equ.~\ref{Nis3_1} is \textbf{true}, When $0<w<1, x_{i}>0, N \in \mathbb{N}$.
\end{description}
Therefore, When we apply \textbf{Lemma 2.} in the loss function $L$, we could got the following relationship.
\begin{equation}\label{weighted_loss_2}
    \begin{aligned}
        L&= -\frac{1}{N} \sum^{N}_{i}\log \left( \frac{e^{w_{i}\cdot f_{y_{i}}}} {(\sum^{K}_{j} e^{f_{j}})^{w_{i}}} \right) \\&
        \leqslant -\frac{1}{N} \sum^{N}_{i}\log \left( \frac{e^{w_{i}\cdot f_{y_{i}}}} {\sum^{K}_{j} e^{w_{i}\cdot f_{j}}} \right)
    \end{aligned}
\end{equation}
From Equ.~\ref{weighted_loss_2}, we could know that we could re-weight the feature map instead of loss value.

\subsection{Experiments with Lightweight Networks}
We present more results with classical lightweight networks (DensNet-100~\cite{densenet}, ShuffleNetv2~\cite{ShunffleNetv2}, MobileNetv2~\cite{MobileNetv2} and EfficientNetB0~\cite{EfficientNet}) on CIFAR-100. Especially, the EfficientnetB0 is obtained by Neural Architecture Search (NAS). Results are shown in  the Table.~\ref{tab:Cifar100-results-small-net_NAS}.
\begin{table}[h]
    \centering
    \resizebox{0.45\textwidth}{!}{%
        \begin{tabular}{lccc}
            \toprule[1.0pt]
            {Architecture} & Params(M) & FLOPs(G) & Err (\%) \\ \midrule
            {DensNet-100}~\cite{densenet} & 0.76 & 0.29 & 21.95 \\
            { {+B$\boldsymbol{A}^2$M} (ours)} & 0.78 & 0.30 &   \textbf{20.63\scriptsize{(-1.32)}}
            \\ \midrule
            {ShuffleNetv2}~\cite{ShunffleNetv2} & 1.3 & 0.05 & 26.8 \\
            {{+B$\boldsymbol{A}^2$M} (ours)} &1.92  & 0.06 &  \textbf{25.49\scriptsize{(-1.31)}}
            \\ \midrule
            {MobileNetv2 }~\cite{MobileNetv2} & 2.26 & 0.07 & 30.17 \\
            {{+B$\boldsymbol{A}^2$M} (ours)} & 2.31 & 0.07 & \textbf{26.77\scriptsize{(-3.40)}}
            \\ \midrule
            {EfficientNetB0}~\cite{EfficientNet} & 2.80 & 0.03 & 36.01 \\
            {{+B$\boldsymbol{A}^2$M} (ours)} & 4.18 & 0.03 & \textbf{33.98\scriptsize{(-2.03)}}
            \\ \bottomrule[1.0pt]
        \end{tabular}%
    }
    \caption{Image classification results on  CIFAR-100 with lightweight  networks and NAS network.}
    \label{tab:Cifar100-results-small-net_NAS}
\end{table}

\subsection{Experiments on CIFAR-100}
We presents more results with other attention mechanisms on CIFAR-100 in Table.~\ref{tab:Cifar100-results-different-networks}.

\begin{table}[h]
    \centering
    \resizebox{0.5\textwidth}{!}{%
        \begin{tabular}{lccc}
            \toprule[1.0pt]
            Architecture & Params(M) & FLOPs(G) &  Err (\%) \\ \midrule
            ResNet-50~\cite{ResNet} & 23.71 & 1.22 & 21.49 \\
            +SE~\cite{SE} & 26.24 & 1.23 & 20.72 \\
            +CBAM~\cite{CBAM} & 27.44 & 1.34 & 21.01 \\
            +BAM ~\cite{BAM} & 24.07 & 1.25 & 20.00 \\
            +B$\boldsymbol{A}^2$M (ours) &24.21  & 1.33  &  \textbf{17.60\scriptsize{(-3.89)}}

            \\ \midrule

            ResNet-101~\cite{ResNet} & 42.70 & 2.44 & 20.00 \\
            +SE~\cite{SE} & 45.56 & 2.54 & 20.89 \\
            {+CBAM}~\cite{CBAM} & 49.84 & 2.54 & 20.30 \\
            {+BAM}~\cite{BAM} & 43.06 & 2.46 & 19.61 \\
            {{+B$\boldsymbol{A}^2$M} (ours)} & 42.32 & 2.55 &\textbf{17.88\scriptsize{(-2.12)}}

            \\ \midrule

            {PreResNet-110}~\cite{Preresnet} & 1.73 & 0.25 & 22.22 \\
            {+BAM}~\cite{BAM} & 1.73 & 0.25 & 21.96 \\
            {+SE}~\cite{SE} & 1.93 & 0.25 & 21.85 \\
            {{+B$\boldsymbol{A}^2$M} (ours)} & 1.86 & 0.25 & \textbf{21.41\scriptsize{(-0.81)}}

            \\ \midrule

            {WRN-28(w=8)}~\cite{wrn} & 23.40 & 3.36 & 20.40 \\
            {+GE}~\cite{GE} & - & - & 19.74 \\
            {+SE}~\cite{SE} & 23.58 & 3.36 & 19.85 \\
            {+BAM}~\cite{BAM} & 23.42 & 3.37 & 19.06 \\
            {{+B$\boldsymbol{A}^2$M} (ours)} & 23.44 & 3.46 & \textbf{17.95\scriptsize{(-2.45)}}

            \\ \midrule

            {WRN-28(w=10)}~\cite{wrn} & 36.54 & 5.24 & 18.89 \\
            {+GE}~\cite{GE} & 36.30 & 5.20 & 20.20 \\
            {+SE}~\cite{SE} & 36.50 & 5.20 & 19.00 \\
            {+BAM}~\cite{BAM} & 36.57 & 5.25 & 18.56 \\
            {+AA}~\cite{AA} & 36.20 & 5.45 & 18.40 \\
            {{+B$\boldsymbol{A}^2$M} (ours)} & 36.59 & 5.39 &  \textbf{17.30\scriptsize{(-1.59)}}

            \\ \midrule

            {ResNeXt-29}~\cite{ResNext} & 34.52 & 4.99 & 18.18 \\
            {+BAM}~\cite{BAM} & 34.61 & 5.00 & 16.71 \\
            {{+B$\boldsymbol{A}^2$M} (ours)} & 34.78 & 5.53 & \textbf{16.10\scriptsize{(-2.08)}}

            \\ \bottomrule[1pt]
        \end{tabular}%
    }
    \caption{Image classification results on CIFAR-100 ~\cite{Cifar100} across a range of ResNet architectures and its variants. The results of compared methods could be find in corresponding papers.}
    \label{tab:Cifar100-results-different-networks}
\end{table}

\subsection{Experiments on ImageNet}\label{imagenet2015}
We presents more results with other attention mechanisms on ImageNet in Table.~\ref{ImageNet-results-different-networks}.

\begin{table}[h]
    \centering
    \resizebox{0.5\textwidth}{!}{%
        \begin{tabular}{lcccc}
            \toprule[1.0pt]
            {Architecture} & Params.(M) & FLOPs(G) & \textbf{Top1Err(\%)} & \textbf{Top5Err(\%)} \\ \midrule
            {ResNet-18~\cite{ResNet}} & 11.69 & 1.81 & 29.60 & 10.55 \\
            {+SE~\cite{SE}} & 11.78 & 1.81 & 29.41 & 10.22 \\
            {+BAM~\cite{BAM}} & 11.71 & 1.82 & 28.88 & 10.10 \\
            {+CBAM~\cite{CBAM}} & 11.78 & 1.82 & 29.27 & 10.09 \\
            {{+B$\boldsymbol{A}^2$M} (ours)} & 12.28 & 1.92 & \textbf{ 28.62\scriptsize{(-0.98)} }& \textbf{ 10.04\scriptsize{(-0.51)} }

            \\ \midrule[1.0pt]

            {ResNet-34~\cite{ResNet}} & 21.80 & 3.66 & 26.69 & 8.60 \\
            {+SE~\cite{SE}} & 21.96 & 3.66 & 26.13 & 8.35 \\
            {+BAM~\cite{BAM}} & - & - & 25.71 & 8.21 \\
            {+CBAM~\cite{CBAM}} & 21.96 & 3.67 & 25.99 & 8.24 \\
            {{+B$\boldsymbol{A}^2$M} (ours)} & 21.92 & 3.77 &  \textbf{25.15\scriptsize{(-1.54)}} & \textbf{8.13\scriptsize{(-0.47)}}

            \\ \midrule[1.0pt]

            {WRN-18(w=1.5)~\cite{wrn}} & 25.88 & 3.87 & 26.85 & 8.88 \\
            {+BAM~\cite{BAM}} & 25.93 & 3.88 & 26.67 & 8.69 \\
            {+SE~\cite{SE}} & 26.07 & 3.87 & 26.21 & 8.47 \\
            {+CBAM~\cite{CBAM}} & 26.08 & 3.87 & 26.10 & 8.43 \\
            {{+B$\boldsymbol{A}^2$M} (ours)} & 28.43& 4.69 &  \textbf{23.60\scriptsize{(-3.25)}} & \textbf{7.60\scriptsize{(-1.28)}}

            \\ \midrule[1.0pt]

            {WRN-18(w=2)~\cite{wrn}} & 45.62 & 6.70 & 25.63 & 8.20 \\
            {+BAM~\cite{BAM}} & 45.71 & 6.72 & 25.00 & 7.81 \\
            {+SE~\cite{SE}} & 45.97 & 6.70 & 24.93 & 7.65 \\
            {+CBAM~\cite{CBAM}} & 45.97 & 6.70 & 24.84 & 7.63 \\
            {{+B$\boldsymbol{A}^2$M} (ours)} & 47.88 & 7.56 &  \textbf{23.60\scriptsize{(-2.03)}}&
            \textbf{7.23\scriptsize{(-0.97)}}

            \\ \midrule[1.0pt]

            {ResNeXt-50~\cite{ResNext}} & 25.03 & 3.77 & 22.85 & 6.48 \\
            {+BAM~\cite{BAM}} & 25.39 & 3.85 & 22.56 & 6.40 \\
            {+SE~\cite{SE}} & 27.56 & 3.77 & 21.91 & 6.04 \\
            {+CBAM~\cite{CBAM}} & 27.56 & 3.77 & 21.92 & 5.91 \\
            {SKNet-50~\cite{SK}} & 27.50 & 4.47 & 20.79 & - \\
            {{+B$\boldsymbol{A}^2$M} (ours)} & 27.25& 4.47 &  \textbf{20.98\scriptsize{(-1.87)}} & \textbf{5.85\scriptsize{(-0.63)}}

            \\ \midrule[1.0pt]

            {ResNeXt-101~\cite{ResNext}} & 44.18 & 7.51 & 21.54 & 5.75 \\
            {+SE~\cite{SE}} & 48.96 & 7.51 & 21.17 & 5.66 \\
            {+CBAM~\cite{CBAM}} & 48.96 & 7.52 & 21.07 & 5.59 \\
            {SKNet-101~\cite{SK}} & 48.90 & 8.46 & 20.19 & - \\
            {{+B$\boldsymbol{A}^2$M} (ours)} & 45.51 & 8.23 & \textbf{ 20.05\scriptsize{(-1.49)}}& \textbf{ 5.22\scriptsize{(-0.53)}}

            \\ \bottomrule[1.0pt]
        \end{tabular}%
    }
    \caption{Image classification results on ImageNet-1K dataset~\cite{ImageNet} across a range of ResNet architectures and its variants.}
    \label{ImageNet-results-different-networks}
\end{table}

\subsection{B$\boldsymbol{A}^2$M in Network}
We perform experiments to determine the choice of B$\boldsymbol{A}^2$M within a network. We use \textbf{four} blocks in the ResNet-50. Thus, we have 15 combinations. The results are shown in Table.~\ref{tab:att_diff_pos_com}. When inserting one B$\boldsymbol{A}^2$M into the ResNet-50, the performance at the second block is the best, and the performance of the first block is the worst. Then, embedding B$\boldsymbol{A}^2$M at the fourth block can put the performance forward further. Furthermore, ResNet-50 has the highest performance ($17.60\%$) when all four blocks are embedded in B$\boldsymbol{A}^2$M. The result reveals the rationality of calculating SARs at variant layers.

\begin{table*}[ht]
    \centering
    \resizebox{0.9\textwidth}{!}{%
        \begin{tabular}{c|c|c|c|c|c|c|c|c|c|c|c|c|c|c|c|c}
            \toprule
            & {No B$\boldsymbol{A}^2$M} & \multicolumn{4}{c|}{{One B$\boldsymbol{A}^2$M}} & \multicolumn{6}{c|}{{Two B$\boldsymbol{A}^2$Ms}} & \multicolumn{4}{c|}{{Three B$\boldsymbol{A}^2$Ms}} & {Four B$\boldsymbol{A}^2$Ms}\\ \hline
            \midrule
            Block1 &  &  \checkmark &  &  &  &  \checkmark &  \checkmark &  \checkmark &  &  &  &  &  \checkmark &  \checkmark &  \checkmark &  \checkmark \\
            Block2 &  &  &  \checkmark &  &  &  \checkmark &  &  &  \checkmark &  \checkmark &  &  \checkmark &  &  \checkmark &  \checkmark &  \checkmark \\
            Block3 &  &  &  &  \checkmark &  &  &  \checkmark &  &  \checkmark &  &  \checkmark &  \checkmark &  \checkmark &  &  \checkmark &  \checkmark \\
            Block4 &  &  &  &  &  \checkmark &  &  &  \checkmark &  &  \checkmark &  \checkmark &  \checkmark &  \checkmark &  \checkmark &  &  \checkmark \\
            \hline
            Mean Err (\%) & 21.49 & 17.96 &  \textbf{17.87} & 17.94 & 17.88 & 18.05 & 18.33 & 17.99 & 17.91 &  \textbf{17.78} & 17.95 &  \textbf{17.70} & 18.02 & 17.90 & 17.93 &  \textbf{17.60} \\
            \vspace{1mm}
            Std & 0.13 & 0.14 & \textbf{0.12} & 0.24 & 0.22 & 0.36 & 0.30 & 0.36 &  \textbf{0.17} & 0.21 & 0.32 &  \textbf{0.11} & 0.18 & 0.17 & 0.21 & 0.15 \\
            \hline
        \end{tabular}%
    }
    \caption{The result of Combination of B$\boldsymbol{A}^2$M within a network. We use ResNet-50 as an instance network. The average error and standard deviation of five random runs are reported and the best results are in  \textbf{bold}.}
    \label{tab:att_diff_pos_com}
\end{table*}

\subsection{Object Detection and Segmentation}\label{Object Detection and segmentation}
In order to further verify the generalization performance of B$\boldsymbol{A}^2$M, we further carried out object detection and instance-level segmentation on the MS COCO \cite{MS_COCO}. The dataset has 80 classes. We train with the union of 118k train images and report ablations on the remaining 5k val images (minival). We train all models for 24 epochs (Lr schd=2x) using synchronized SGD with a weight decay of 0.0001 and a momentum of 0.9. The learning rate is initialized as 0.01 and decays by a factor of 10 at the 16th and 22th epochs. We report bounding box AP of different IoU thresholds form 0.5 to 0.95, and object size (small (S), medium (M), large (L)).We adopt Faster-RCNN \cite{faster_rcnn}and Mask R-CNN framework \cite{mask_rcnn} as our detection method and ImageNet pre-trained ResNet-50/101 as our baseline networks. As shown in Table \ref{tab:COCO_faster_BSE} and Table \ref{tab:COCO_mask_BSE_object_detection}, we can see that the detectors embedded in B$\boldsymbol{A}^2$M are better than the baseline method under different evaluation indicators.

For the instance-level segmentation experiments, we used the classic Mask R-CNN framework \cite{mask_rcnn} as our instance-level segmentation method and ImageNet pre-trained ResNet-50/101 as baseline networks. We report the standard COCO metrics, including AP (averaged over IoU thresholds), $AP_{50}$, $AP_{75}$, and $AP_{S}$, $AP_{M}$, $AP_{L}$ (AP at different scales). Unless noted, AP is evaluating using mask IoU. The results are shown in Table \ref{tab:COCO_mask_BSE}. We can see that regardless of how the evaluation indicators change, the Mask R-CNN embedded in B$\boldsymbol{A}^2$M is even much better than the baseline method. We can conclude that B$\boldsymbol{A}^2$M can boost the performance of the detector based on the improvement of the ability in feature extraction under the condition of the small size of the batch.

For experiments of object detection and instance segmentation, we utilize MMDetection \cite{mmdetection} as the platform. All the related hyper-parameters are not changed. Here we study the performance improvement after inserting B$\boldsymbol{A}^2$M into the backbone network. Because we use the same detection method, the performance gain can only be attributed to the feature enhancement capabilities, given by B$\boldsymbol{A}^2$M.

\begin{table*}
    \setlength{\abovecaptionskip}{0.cm}
    \setlength{\belowcaptionskip}{-0.cm}
    \resizebox{\textwidth}{!}{%
        \begin{tabular}{llcccccc}
            \toprule
            Detector  & backbone & $AP^{bb}$ & $AP^{bb}_{50}$ & $AP^{bb}_{75}$ & $AP^{bb}_{S}$ & $AP^{bb}_{M}$ & $AP^{bb}_{L}$ \\ \midrule
            Faster R-CNN  & ResNet-50& 37.60  & 58.90 & 40.90 & 21.80 & 41.50 & 48.80  \\
            Faster R-CNN & ResNet-50 w B$\boldsymbol{A}^2$M & \textbf{38.50\scriptsize{(+0.9)}} & \textbf{60.00\scriptsize{(+1.1)}} & \textbf{41.90 \scriptsize{(+1.0)}}  & \textbf{22.20 \scriptsize{(+0.4)}} & \textbf{42.10\scriptsize{(+0.6)}} & \textbf{49.80\scriptsize{(+1.0)}} \\\bottomrule
            Faster R-CNN  & ResNet-101& 39.40  & 60.60 & 43.00 & 22.10 & 43.60 & 52.10  \\
            Faster R-CNN & ResNet-101 w B$\boldsymbol{A}^2$M & \textbf{40.60\scriptsize{(+1.2)}} & \textbf{62.10\scriptsize{(+1.5)}} & \textbf{44.40 \scriptsize{(+1.4)}}  & \textbf{22.70 \scriptsize{(+0.6)}} & \textbf{44.30\scriptsize{(+0.7)}} & \textbf{53.60\scriptsize{(+1.5)}} \\ \bottomrule
        \end{tabular}%
    }
    \caption{\textbf{Object detection} single-model results on COCO minival (bounding box AP).The baseline detector is \textbf{ Faster R-CNN} with ResNet-50/101 .}
    \label{tab:COCO_faster_BSE}
\end{table*}

\begin{table*}
    \setlength{\abovecaptionskip}{0.cm}
    \setlength{\belowcaptionskip}{-0.cm}
    \centering
    \resizebox{\textwidth}{!}{%
        \begin{tabular}{llcccccc}
            \toprule
            Detector  & backbone  & $AP^{bb}$ & $AP^{bb}_{50}$ & $AP^{bb}_{75}$ & $AP^{bb}_{S}$ & $AP^{bb}_{M}$ & $AP^{bb}_{L}$ \\ \midrule
            Mask R-CNN & ResNet-50 & 38.30 & 59.70 & 41.30 & 22.50 & 41.70 & 50.40  \\
            Mask R-CNN  & ResNet-50 w B$\boldsymbol{A}^2$M & \textbf{39.80\scriptsize{(+1.50)}} & \textbf{60.90\scriptsize{(+1.20)}} & \textbf{43.70\scriptsize{(+2.40)}} & \textbf{23.00\scriptsize{(+0.50)}} & \textbf{43.10\scriptsize{(+1.40)}} & \textbf{52.10\scriptsize{(+1.70)}} \\\bottomrule
            Mask R-CNN  & ResNet-101 & 40.40 &61.50 &44.10 &22.20 & 44.80& 52.90  \\

            Mask R-CNN  & ResNet-101 w B$\boldsymbol{A}^2$M & \textbf{42.20\scriptsize{(+1.80)}} & \textbf{62.90\scriptsize{(+1.40)}} & \textbf{46.60\scriptsize{(+2.50)}} & \textbf{24.00\scriptsize{(+1.80)}} & \textbf{46.70\scriptsize{(+1.90)}} & \textbf{55.30\scriptsize{(+2.40)}} \\ \bottomrule
        \end{tabular}%
    }
    \caption{\textbf{Object detection} single model results on COCO minival (bounding box AP). The baseline is \textbf{Mask R-CNN} with ResNet-50/101.}
    \label{tab:COCO_mask_BSE_object_detection}
\end{table*}

\begin{table*}
    \centering
    \resizebox{\textwidth}{!}{%
        \begin{tabular}{llcccccc}
            \toprule
            Method  & backbone & AP & $AP_{50}$ & $AP_{75}$ & $AP_{S}$ & $AP_{M}$ & $AP_{L}$ \\ \midrule
            Mask R-CNN  & ResNet-50 &34.80  & 56.10  & 36.90  & 16.10  & 37.20  & 52.90  \\
            Mask R-CNN  & ResNet-50 w B$\boldsymbol{A}^2$M & \textbf{36.10\scriptsize{(+1.30)}} & \textbf{57.80\scriptsize{(+1.70)}} & \textbf{38.80\scriptsize{(+1.90)}} & \textbf{17.00\scriptsize{(+0.90)}} & \textbf{38.70\scriptsize{(+1.50)}} & \textbf{52.90\scriptsize{(+0.00)}} \\\bottomrule
            Mask R-CNN  & ResNet-101 &36.50  & 58.10  & 39.10  & 18.40  & 40.20  & 50.40  \\
            Mask R-CNN  & ResNet-101 w B$\boldsymbol{A}^2$M & \textbf{37.70\scriptsize{(+1.20)}} & \textbf{59.70\scriptsize{(+1.60)}} & \textbf{40.90\scriptsize{(+1.80)}} & \textbf{18.60\scriptsize{(+0.20)}} & \textbf{41.00\scriptsize{(+0.80)}} & \textbf{55.90\scriptsize{(+5.50)}} \\ \bottomrule
        \end{tabular}%
    }
    \caption{\textbf{Instance segmentation} single-model results on COCO minival (mask AP). The  baseline is \textbf{Mask R-CNN} with  ResNet-50/101.}
    \label{tab:COCO_mask_BSE}
\end{table*}


\ifCLASSOPTIONcaptionsoff
  \newpage
\fi



\bibliographystyle{IEEEtran}
\bibliography{egbib}
%

\end{document}